\documentclass[twoside,11pt]{article} 
\usepackage[letterpaper, margin=1in]{geometry}

\usepackage{MnSymbol}
\usepackage{hyperref}
\usepackage{url}
\usepackage{setspace}
\usepackage{amsthm}
\usepackage{bm}
\usepackage{xcolor}      
\usepackage{framed}
\usepackage{pdfpages}
\usepackage{graphicx}
\usepackage{float}
\usepackage{bbold,color}
\usepackage[authoryear]{natbib} 
\newsavebox{\measurebox}
\usepackage{parskip}
\usepackage{algorithm}
\usepackage{algpseudocode}
\usepackage[normalem]{ulem}
\usepackage{multirow}
\usepackage{cancel}
\usepackage{booktabs}
\usepackage{natbib}
\usepackage{amsmath}
\usepackage{bbm}
\usepackage{float}
\usepackage{ulem}
\usepackage{preamble}
\usepackage{subcaption}
\usepackage{multirow}

\newtheorem{remark}{Remark}[section]
\title{Deconfounded Warm-Start Thompson Sampling with Applications to Precision Medicine
}

\author{%
  Prateek Jaiswal\thanks{Daniels School of Business, Purdue University. jaiswalp@purdue.edu}%
  \and
  Esmaeil Keyvanshokooh\thanks{Mays Business School, Texas A\&M University. keyvan@tamu.edu}%
  \and
  Junyu Cao\thanks{McCombs School of Business, University of Texas at Austin. junyu.cao@mccombs.utexas.edu}%
}
\date{}

\begin{document}

\maketitle

\begin{abstract}
  Randomized clinical trials often require large patient cohorts before drawing definitive conclusions, yet abundant observational data from parallel studies remains underutilized due to confounding and hidden biases. To bridge this gap, we propose Deconfounded Warm-Start Thompson Sampling (DWTS), a practical approach that leverages a Doubly Debiased LASSO (DDL) procedure to identify a sparse set of reliable measured covariates and combines them with key hidden covariates to form a reduced context. By initializing Thompson Sampling (LinTS) priors with DDL-estimated means and variances on these measured features—while keeping uninformative priors on hidden features—DWTS effectively harnesses confounded observational data to kick-start adaptive clinical trials. Evaluated on both a purely synthetic environment and a virtual environment created using real cardiovascular risk dataset, DWTS consistently achieves lower cumulative regret than standard LinTS, showing how offline causal insights from observational data can improve trial efficiency and support more personalized treatment decisions.
\end{abstract}

\section{Introduction}

In many sequential decision-making problems, the decision-maker has access to contextual information that can guide personalized actions. Contextual Multi-Armed Bandit (CMAB) algorithms formalize this setting, extending traditional multi-armed bandits by incorporating side information (e.g., user attributes, environmental factors, or historical states) to inform arm selection \citep{lattimore2020bandit}. This contextual adaptability makes CMABs especially valuable in dynamic, high-stakes applications like personalized healthcare, targeted advertising, and adaptive pricing. 

A growing body of literature explores leveraging  \textit{observational (offline) data} (collected from past interactions without active experimentation) to initialize and accelerate learning in CMABs \citep{tennenholtz2021bandits, hao2023leveraging}. Offline data provides a valuable source of information on context–arm–outcome relationships, which can substantially reduce the need for costly, risky, or even unethical online exploration in high-stakes settings. 

More importantly, observational data can play a critical role in identifying a sparse set of \textit{informative features} that are most predictive of treatment responses. In fact, in high-dimensional settings, not all features equally influence the effectiveness of an action. Many features may be irrelevant or weakly correlated with the outcome, introducing noise and exacerbating the sample efficiency of online learning algorithms. By exploiting the advantages of observational data, we can perform feature selection or dimensionality reduction to isolate the subset of covariates with high predictive power for treatment outcomes. This approach enables more targeted data collection efforts during online exploration, focusing resources on the most impactful variables, and thereby further lowering the sample and cost burden of sequential experimentation.

However, the offline observational datasets often suffer from \textit{confounding bias}: unobserved variables that simultaneously influence both the action selection and the resulting outcome. Failing to adjust for such confounders leads to biased policy learning, which in turn results in suboptimal or even harmful decisions, especially in high-stakes applications such as personalized medicine. Therefore, while observational data holds promise for improving sample efficiency and guiding exploration via sparsity, its effective use demands rigorous methods for confounding adjustment and debiasing to ensure reliable and robust decision-making.

The integration of observational and experimental data is therefore not only just a methodological improvement, but also a fundamental necessity in modern healthcare decision-makings. As medicine increasingly embraces personalization, the limitations of relying solely on either observational data or randomized controlled trials (RCTs) become clearly apparent. \textit{Observational data}, collected through electronic health records (EHRs), registries, and routine clinical care, offers unparalleled access to large-scale, real-world patient populations, capturing a breadth of demographics, comorbidities, and care pathways. However, its non-randomized nature inherently introduces confounding biases, which complicates causal decision-makings and inferences \citep{hatt2022combining, stuart2011use}. Conversely, \textit{experimental data} (RCTs) remain the gold standard for causal estimation but are limited by strict eligibility criteria, artificially controlled settings, and high costs, restricting their scalability and generalizability. These complementary strengths and weaknesses highlight a critical gap: neither data source alone suffices for developing robust, scalable, and individualized decision-making systems in healthcare. RCTs inform us of average treatment effects, but fail to reflect the rich heterogeneity of real-world patient populations. Observational data reveals this heterogeneity, yet lacks the causal clarity to confidently guide interventions. Bridging these paradigms is essential to develop learning algorithms that are both scientifically rigorous and practically applicable.

A motivating example can be seen in the critical problem of medical treatment selections \citep{denton2018optimization, keyvanshokooh2025contextual}, where clinicians must choose between medication and physical therapy for chronic disease patients. Observational historical data may show that those with severe symptoms typically receive more aggressive medication and experience poorer outcomes \citep{stukel2007analysis}. If clinicians fails to adjust for the severity of the illness, it might incorrectly conclude the medication less effective \citep{norgaard2017confounding, dahabreh2024causal}. This example highlights that neglecting such confounders can lead to flawed assessments of treatment efficacy \citep{brookhart2010confounding, powell2022exploration}, underscoring the critical need for developing new methodologies to address these biases effectively. Accordingly, integrating observational and experimental data enables more nuanced and patient-centered interventions. As the healthcare landscape shifts toward data-driven precision medicine, the synergistic integration of observational and experimental data becomes critical. 
Our work directly addresses this pressing need by proposing methodologies that integrate these different data sources to enable robust, scalable, and personalized sequential decision-making in high-stakes domains like healthcare.

Specifically, this paper addresses this critical challenge by making the following contributions. (i) We propose a unified sequential decision‑making framework that leverages a structural equation model (SEM) to characterize the observational data‑generation process -- explicitly modeling hidden confounders -- and couples it with an online environment modeled through a linear response model where those previously unobserved factors are directly measured. (ii) Building on this framework, we introduce Deconfounded Warm‑Start Thompson Sampling (DWTS), which uses a Doubly Debiased LASSO (DDL)~\citep{guo2022doubly} to select a sparse set of measured covariates, augments them with the hidden features to form a reduced context, and then warm‑starts a Linear Thompson Sampling (LinTS) algorithm~\citep{agrawal2013thompson} by initializing its priors on the selected measured dimensions with DDL‑estimated means and variances (while leaving priors on hidden features uninformative). (iii) We evaluate DWTS both on a purely synthetic environment and a virtual environment built from real cardiovascular‐risk dataset. In each case, we benchmark against standard LinTS baselines as well as the state‐of‐the‐art OFUL algorithm for partially observable confounded data~\citep{tennenholtz2021bandits}. Across all comparisons, DWTS attains significantly lower cumulative regret and rapidly converges to more effective treatment policies.

\subsection{Literature Review}

\textbf{Estimating heterogeneous treatment effects using observational data}.
Estimating heterogeneous treatment effects is of great importance and many machine learning algorithms have been adapted to address this challenge \citep{kunzel2019metalearners,hu2021estimating}. For example, random forest-based methods \citep{wager2018estimation} and deep learning algorithms \citep{hatt2021estimating,shi2019adapting,yao2018representation,curth2021nonparametric} have been recently studied. However, one major challenge in making causal inferences from observational data is the presence of hidden confounders. These are factors that impact both the assignment of treatment and the outcome, yet remain unmeasured in the observational dataset. It has been shown that one can never prove that there is no hidden confounding in an observational study \citep{pearl2009causality}.

Another area of research focuses on using latent variable models to uncover hidden confounding variables. For example, \cite{kuzmanovic2021deconfounding} develop a method that leverages observed noisy proxies to learn a hidden embedding that reflects the true hidden confounders. Other methods are based on variational autoencoders \citep{louizos2017causal} and factor models \citep{wang2019blessings}. However, there remains uncertainty regarding whether the observed covariates accurately represent proxies for the true confounding factors.

\textbf{Combining observational and randomized data} Some recent work focuses on combining observational and randomized data. For example, \cite{kallus2018removing} introduce a novel method of using limited experimental data to mitigate hidden confounding in causal effect models trained on larger observational datasets. Some recent studies suggest methods involving the estimation of two distinct estimators \citep{hatt2022combining,rosenman2023combining,yang2019combining,cheng2021adaptive}: one applied to observational data, prone to bias, and another applied to randomized data, free from bias. \cite{ilse2021combining} introduce a method for integrating interventional and observational data through causal reduction. While previous studies have concentrated on utilizing a \emph{static} observational and randomized dataset, our emphasis lies on addressing a sequential decision-making scenario. Here, the primary objective is to optimize the cumulative reward of online decisions by leveraging a large observational dataset.

\textbf{Warm-starting learning algorithms} A few studies focus on using labeled dataset to warm-start the online learning system \citep{banerjee2022artificial}. For example, \cite{tennenholtz2021bandits}  study linear contextual bandits with access to a partially observable confounded dataset, where only some features are observable. By reducing the partially observable data to linear constraints, an OFUL (optimism-in-the-face-of-uncertainty) algorithm is proposed. \cite{sharma2020under}  study a variant of the MAB problem where side information in the form of bounds on the mean of each arm is provided.  
Some other literature attempts to bridge the gap between online and offline reinforcement learning. For instance, \cite{wagenmaker2023leveraging} proposes an algorithm that has access to an offline dataset but can also augment this for fine-tuning via online interactions with the environment. \cite{tang2023efficient} proposes a Bayesian approach for online reinforcement learning in the infinite horizon setting when there is an offline dataset to start with.  However, the above-mentioned literature assumes that both offline and online data are generated from the same underlying model. 

\cite{zhang2019warm} consider settings where the underlying learning signal may differ between online and offline data sources and investigate the feasibility of learning from a mix of both fully-labeled supervised data and contextual bandit data. However, while they assume that the context distribution is consistent between the online and offline data, we account for a more practical confounding bias present in the offline dataset.

\textbf{Bandits with partially observable features.} Recent work by~\cite{kim2025linear} introduces a doubly robust OFUL algorithm tailored to settings where certain covariates remain unobserved throughout the online interaction. They address this challenge by augmenting the observed feature space with orthogonal basis vectors and employing a doubly robust estimator to achieve sublinear regret without prior knowledge of the unobserved feature space.  In contrast, our setting assumes that while some features are unobserved in the offline data, all features become fully observable during the online phase. 

\textbf{Variable selection algorithms} Variable selection is a critical process in machine learning that involves identifying the most relevant features or variables from a dataset to improve model performance. High-dimensional statistical problems emerge across a wide range of scientific disciplines and technological applications, including healthcare \citep{schneeweiss2017variable} and economics \citep{belloni2013inference,mei2024lasso}, etc. Several methods have been developed for high dimensional linear regression such as the lasso \citep{tibshirani1996regression}, boosting \citep{buehlmann2006boosting}, and Least angle regression \citep{efron2004least}. More recently, a variant of the Lasso algorithm has been developed \citep{yamada2014high,kim2019hi,luo2014sequential}. In this work, we apply the Doubly Debiased Lasso (DDL) method \citep{guo2022doubly} to an offline dataset for high-dimensional feature selection.

\section{Modeling Framework}

We model the sequential decision-making problem using a contextual bandit framework, where we aim to combine the offline observational data with the online data collected while interacting with the environment to make more informed decisions efficiently. 
We denote a collection of $n$ samples of observational data as $\cD^o := \{(Z_i^o, A_i^o, Y_i^o)\}_{i=1}^n$ and represent the online data collected up to time $t$ as $\cD_t^h := \{(X_s^h, A_s^h, Y_s^h)\}_{s=1}^{t}$, where $Z_i^o \in \R^p$ and $X_i^h \in \cX \subseteq \R^{p+q}$ denote high-dimensional (but sparse) covariates or contexts. Here, $A_i^{(\cdot)} \in \cA := [K]$ is the treatment, and $Y_i^{(\cdot)} \in [0,1]$ is the outcome. The first $p$ components of $X_{(\cdot)}^h$ correspond to the same features observed in $Z_{(\cdot)}^o$. Because observational data is typically confounded, we denote the (non-sparse) hidden confounders for the observational data as $H_i^o \in \R^q$. We also assume that in the online experimental setting, the corresponding confounders $H_i^h \in \R^q$ are observed/measured as the final $q$ components of $X_i^h$. For any $N \in \mathbb{N}$, we denote $[N] := \{1,2,\ldots,N\}$. For any $x \in \R^D$ and $A \in \R^{D \times D}$, we define the matrix norm as $\|x\|_A = \sqrt{x^\top A x}$, and use $\|\cdot\|$ to denote the Euclidean norm. We assume that the observational covariates $\{Z_i^o\}$ are fixed (non-random), while the online covariates $\{X_t^h\}$ are chosen randomly, which could be adversarial. 

At each time $t \in [T]$, the decision-maker (DM) observes a context $X_t^h \in \cX$. After observing the context, the DM selects a treatment $A_t \in \cA$ using all past data, including both observational and online data, i.e., $\cD_{t-1}^h \cup \cD^o$. Subsequently, the DM observes a reward $Y_t^h(A_t^h) \in [0,1]$ corresponding to the selected action. We assume that $Y_t^h(a)$ for each $a \in [K]$ is generated independently of $A_t$, but only $Y_t^h(A_t)$ is revealed.

\paragraph{Observational Data Model:}
 We assume the existence of a true but unknown set of modeling parameters, denoted as $\bmtheta^* = [\theta_1^*, \theta_2^*, \ldots, \theta_K^*]$, where each $\theta_a^* \in \R^p$ for $a \in [K]$. The outcome $Y_j^o$ under treatment $a \in [K]$ in the observational data is generated according to the following linear Structural Equation Model (SEM). For each $a \in [K]$ and $Z_j^o \in \R^p$, we have:
\begin{align}
    Y_j^o &= {\theta_a^*}^\top Z_j^o + {\phi_a^*}^\top H_j^o + \epsilon_j, \quad \text{and} \quad 
    Z_j^o = {\Psi_a^*}^\top H_j^o + E_j, \quad \text{for } 1 \leq j \leq n_a, \label{eq:Offline}
\end{align}
where $n_a$ denotes the number of observational samples with treatment $A_i^o = a$. The noise $\epsilon_j \sim \cN(0,1)$ is independent of $Z_j^o \in \R^p$, $H_j^o \in \R^q$, and $E_j \in \R^p$. Moreover, the components of $E_j$ are uncorrelated with those of $H_j^o$. The parameters $\phi_a^* \in \R^q$ and $\Psi_a^* \in \R^{p \times q}$ capture the hidden confounding structure in the observational data. We collect these into $\bm \phi^* = [\phi_1^*, \ldots, \phi_K^*]$ and $\bm \Psi^* = [\Psi_1^*, \ldots, \Psi_K^*]$. 

In healthcare applications, the hidden confounders $H_j^o$ may represent unobserved demographic or clinical factors present in the observational records, such as patient socioeconomic status, genetic predispositions, or prior health conditions that are not systematically recorded. These factors can simultaneously influence both the assigned treatment $A_j^o$ and the outcome $Y_j^o$, thereby inducing spurious correlations if not properly accounted for. Modeling the relationships between the observed covariates $Z_j^o$ and the hidden confounders $H_j^o$  allows us to capture and mitigate such confounding effects, which is critical for obtaining reliable estimates of the causal effect parameters $\theta_a^*$.

In personalized recommendation platforms, the hidden confounders \(H_j^o\) may encode unobserved user traits or situational factors—such as transient interests, browsing mood, or social influences—that are not directly captured in the logged data. For instance, a user’s fleeting interest in travel  might simultaneously increase the likelihood of being shown vacation‐related content (\(A_j^o\)) and clicking on those recommendations (\(Y_j^o\)). 

In our model, we allow the observed features \(Z_j^o\) (e.g.\ past click counts, time of day) to partially reflect these hidden factors, and thus adjust our estimate of the true recommendation effect \(\theta_a^*\) for such confounding.

\paragraph{Online Data Model:}

Since, in the online setting, the features that were hidden in the observational data are now observed as part of $X_t^h$, we model the online data generation process using a linear model that shares the same parameters $\bm\theta^*$ and $\bm\phi^*$. For each $a \in [K]$ and $X_t^h \in \cX$, the outcome under action $a$ is given by:
\begin{align}
    Y_t^h(a) = [\theta_a^*, \phi_a^*]^\top X_t^h + \epsilon_t, \label{eq:Online}
\end{align}
where the noise term $\epsilon_t \sim \cN(0,1)$ is independent of the covariates $X_t^h$.

Different from the observational data model, in the online  setting, we assume that all relevant features are observed, and there are no hidden confounders. This assumption is often justified in modern applications where platforms have the ability to design and control the data collection process. 

For example, in an e-commerce platform, the context \(X_j^h\) may include detailed user activity signals such as recent browsing history, time spent on different product categories, cart status, device information, and geolocation. These features are directly measured at the time of decision making and are designed to capture the user’s immediate purchase intent and preferences. Because the platform engineers the system to log these comprehensive features, there is little risk of omitted variables influencing both the action selection \(A_j^t\) and the outcome \(Y_j^t\).

Similarly, in online news recommendation, the context may include the user's recent reading history, time of day, current session activity (e.g., number of articles viewed, scroll behavior), and even explicit feedback such as likes or shares. Since these observable features fully characterize the user's current interests and engagement state, it is reasonable to assume that the decision-making is based entirely on \(X_j^t\), without hidden confounding.

We introduce a two-step meta-algorithm that leverages confounded offline data to initialize and accelerate online learning. In the first step, we apply off-the-shelf debiasing methods DDL to the offline dataset in order to obtain an estimate $\hat{\theta}_a$ for each arm $a \in [K]$. In the second step, we deploy LinTS to learn only important variables of $\theta_a^*$ with $\phi^*_a$. Specifically, during the online phase, we warm-start the LinTS algorithm by fixing the prior mean and variance for only the selected (using some threshold) coordinates of $\theta^*_a$ using the estimates from the offline stage. The prior for learning $\phi^*_a$ is set to standard Gaussian. In addition to warm starting the online step, LinTS is learning only $p_{\text{eff}} + q$ dimensions, instead of the full $p + q$ dimensions, making the learning process more efficient (under the sparsity assumptions) than learning from scratch.

\section{Deconfounded Warm-start Thompson Sampling (DWTS)}

In this section, we propose an algorithm that integrates an offline estimation procedure with an online sequential learning process. In the offline phase, for each action $a \in [K]$, the algorithm employs the DDL method~\citep{guo2022doubly} to compute an estimate $\hat{\theta}_{a,i}$ and its associated standard error $\hat{\sigma}_{a,i}$ for each coordinate $i \in [p]$ of the true parameter $\theta^*_{a}$. To identify the important coordinates, we apply a thresholding rule using a hyperparameter $\kappa_o$ (chosen close to $0$). Specifically, we construct a binary masking vector $\hat{m}_a = \left\{ \mathbb{I}(|\hat{\theta}_{a,i}| \geq \kappa_o), \forall i \in [p] \right\}$ to indicate the selected variables. We denote the corresponding selected vectors by adding a superscript $\eff$, for instance, $\theta^{*,\eff}_a = \theta^*_a[\hat m_a] \in \R^{p_{\eff}}$. It is possible that for each $a\in[K]$, we have different $\hat m_a$, but we chose to do not overburden the notations for brevity as it would be clear from the equation which action we are referring to.

Prior to starting the online phase, we fix the prior on the unknown parameter $\theta^{*,\textit{eff}}_a$ and $\phi^*_a$ as follows: We assume that the posterior follows a Gaussian distribution with mean $\hat \mu_{t,a} \in \R^{p_{\eff}+q}$ and precision matrix $\hat B_{t,a}\in \R^{(p_{\eff}+q)\times (p_{\eff}+q)}$. We warm start online learning to learn $\pmb \theta^{*,\eff}$ and $\pmb \phi^*$ by fixing $\hat \mu_{0,a}=[\hat \theta_a^{\eff},\mathbf{0}_q]$ and $\hat B_{0,a}^{-1}= \text{diag}([\hat\sigma_a^{\eff},\mathbf{1}_q])$, where  $\mathbf{0}_q$ and $\mathbf{1}_q$ are a $q$-dimensional vectors of zeros and ones.  During the online phase, at each time step $t\in[T]$, after observing the covariates $X_t^h$ at time $t$, the online learner explores/exploit the environment by selecting an action $A_t^h$. To select $A_t^h$, the online learner first samples a mean vector for each $a\in[K]$ from $\mu_{t,a} \sim \cN (\hat \mu_{t-1,a},\hat B_{t-1,a}^{-1})$, where \[\hat \mu_{t-1,a}= \hat B_{t-1,a}^{-1} \Big( \hat B_{0,a} \hat \mu_{0,a} + \sum_{s=1,A_s^h=a}^{t-1} X_s^{h,\eff} Y_s^h(a) \Big ) \] and \[B_{t-1,a}= \Big( \hat B_{0,a} + \sum_{s=1,A_s^h=a}^{t-1} (X_s^{h,\eff})^\top X_s^{h,\eff} \Big ).\] Then, it selects $A_t^h$ as $a\in[K]$ that maximizes  $\mu_{t,a}^\top X_t^h$. In response to the learner's action at time $t$, the environment provides a feedback (or outcome) $Y_t^h$. Thereafter, the learner uses this feedback and compute $\hat \mu_{t,A_t^h} =\hat B_{t,A_t^h}^{-1} ( \hat B_{t-1,A_t^h} \hat \mu_{t-1,A_t^h} + X_t^{h,\eff} Y_t^h)$ and $\hat B_{t,a} = \hat B_{t-1,a} + (X_t^{h,\eff})^\top X_t^{h,\eff} $.

Effectively, in the online step, LinTS is learning only $p_{\text{eff}} + q$ dimensions instead of the full $p + q$ dimensions, making the learning process more efficient under sparsity assumptions than learning from scratch.
%After observing the covariates $X_t^h$ at time $t$, the online learner $\Pi_h$ explores the environment by selecting an action $A_t^h$, where $A_t^h$ is randomly sampled from $\Pi_h(\cdot|X_t^h, \hat\bmmu_{t-1}\hat{\bmB}_n, D_{t-1}^h)$. 
The pseudo-code for the algorithm is presented in Algorithm~\ref{alg:OOFS}.
\begin{algorithm}
    \caption{Deconfounded Warm-start Thompson Sampling (DWTS)}\label{alg:OOFS}
    \begin{algorithmic}[1]
    \Require Offline dataset $\mathcal{D}^{o}$, threshold $\kappa_o$, time horizon $T$, number of actions $K$
    \smallskip
    \State \textbf{Offline Phase:}
    \For{action $a \in [K]$}
        \State Estimate $\hat{\theta}_{a,i}$ and $\hat{\sigma}_{a,i}$ using $DDL(\cD^o)$ for all $i \in [p]$
        \State Create mask $\hat{m}_a = \left\{ \mathbb{I}(|\hat{\theta}_{a,i}| \geq \kappa_o), \forall i \in [p] \right\}$
        \State Define $\hat{\theta}^{\eff}_a = \hat{\theta}_a[\hat{m}_a]$, $\hat{\sigma}^{\eff}_a = \hat{\sigma}_a[\hat{m}_a]$
        \State Initialize prior mean: $\hat{\mu}_{0,a} = [\hat{\theta}_a^{\eff}, \mathbf{0}_q]$
        \State Initialize prior precision: $\hat{B}_{0,a}^{-1} = \text{diag}([\hat{\sigma}_a^{\eff}, \mathbf{1}_q])$
    \EndFor
    \State Set $\cD_0^h = \{\}$
    \smallskip
    \State \textbf{Online Phase:}
    \For{$t = 1, 2, \ldots, T$}
        \State Observe covariate $X_t^h$
        \For{action $a \in [K]$}
            \State Sample $\mu_{t,a} \sim \cN(\hat{\mu}_{t-1,a}, \hat{B}_{t-1,a}^{-1})$
            \State Compute score: $s_{t,a} = \mu_{t,a}^\top X_t^h$
        \EndFor
        \State Sample action $A_t^h = \arg\max_{a \in [K]} s_{t,a}$
        \State Observe $Y_t^h$ for action $A_t^h$
        \State Set $\cD_t^h = \cD_{t-1}^h \cup \{X_t^h, A_t^h, Y_t^h\}$

        \State \textbf{Update posterior for $A_t^h$:}
        \State $\hat{B}_{t,A_t^h} = \hat{B}_{t-1,A_t^h} + (X_t^{h,\eff})^\top X_t^{h,\eff}$
        \State $\hat{\mu}_{t,A_t^h} = \hat{B}_{t,A_t^h}^{-1} \left( \hat{B}_{t-1,A_t^h} \hat{\mu}_{t-1,A_t^h} + X_t^{h,\eff} Y_t^h \right)$
    \EndFor
    \end{algorithmic}
\end{algorithm}
\begin{remark}[The choice of $\kappa_o$]~\label{rem:kappa}
In Algorithm \ref{alg:OOFS}, the parameter $\kappa_o$ can be selected as a tuning parameter based on the estimates computed in the step 3 of~Algorithm~\ref{alg:OOFS}. Alternatively, it can be determined using theoretical considerations. A confidence interval with asymptotic coverage $(1-\alpha)$ has been established in (13) in \cite{guo2022doubly}. Having this confidence interval, denoted as $CI(\theta_{a,i})=(\hat{\theta}_{a,i}-z_{1-\alpha/2}\hat\sigma_{a,i},\hat{\theta}_{a,i}+z_{1-\alpha/2}\hat\sigma_{a,i})$, we can choose $\kappa_o=\min\{\beta_{\Theta}-\max_{a,i}\{z_{1-\alpha/2}\hat\sigma_{a,i}\},0\}$ where $\beta_{\Theta}=\min_{a,i}\{|\theta_{a,i}|:\theta_{a,i}\neq 0\}$. However, note that the theoretical approach require the knowledge of the minimum of the true important coefficients, which is infeasible in real world experimental settings.
\end{remark}

\section{Synthetic Experiments}

To validate DWTS in a controlled setting, we conduct a set of synthetic experiments. We consider a high-dimensional setup with $K=2$ arms, observed covariate dimension $p=\{20,40,100\}$ (with effective dimension $p_{\text{eff}}=5$), and $q=3$ hidden confounders. For each arm $a \in \{1, 2\}$, we define
\[
\theta^{*}_a = \bigl[(a+1)\,\mathbf{1}_5;\;\mathbf{0}_5\bigr], \quad
\phi^*_a \sim \mathcal{N}(0, I_3).
\]

For each $a \in [K]$, we generate an offline dataset of size $n=1000$ using the observational data model~(1), with $\Psi^*_a$ sampled from $\mathcal{N}(0, \mathbf{1}_{q \times p})$. We then estimate $\hat{\theta}_{a,i}$ and $\hat{\sigma}_{a,i}$ for all $i \in [p]$ using DDL algorithm. Since, this a synthetic experiment, we compute $\kappa_o$ using the strategy describe in~Remark~\ref{rem:kappa} for $\alpha=0.05$. After, computing $\kappa_o$, we compute the mask $\hat{m}_a$ and derive $\hat{\theta}^{\text{eff}}_a$ and $\hat{\sigma}^{\text{eff}}_a$. Each mask $\hat{m}_a$ identifies estimated important measured covariates for arm $a$.

Using the same true sparse weights in the online environment, we compare LinTS, DWTS, and OFUL with partially observable offline data algorithm (black) proposed in~\cite{tennenholtz2021bandits}. In LinTS, we learn all $=p+q$ dimensions from scratch without using offline data. In DWTS, as described in the algorithm, we warm-start the online learning process using only the important features flagged in the offline phase, along with the confounders ($=p_{\eff}+q$). At each round $t$, both algorithms observe $x_t \in \mathbb{R}^{p+q}$, select an arm $A_t^h$, and receive a reward $ Y_t^h(A_t^h)$. Both methods then update their posterior distributions. All experiments are conducted over a horizon of $T = 1000$ rounds and repeated across $50$ independent sample paths.

\begin{figure}[t]
  \centering
  \begin{subfigure}{0.32\linewidth}
    \centering
    \includegraphics[width=\linewidth]{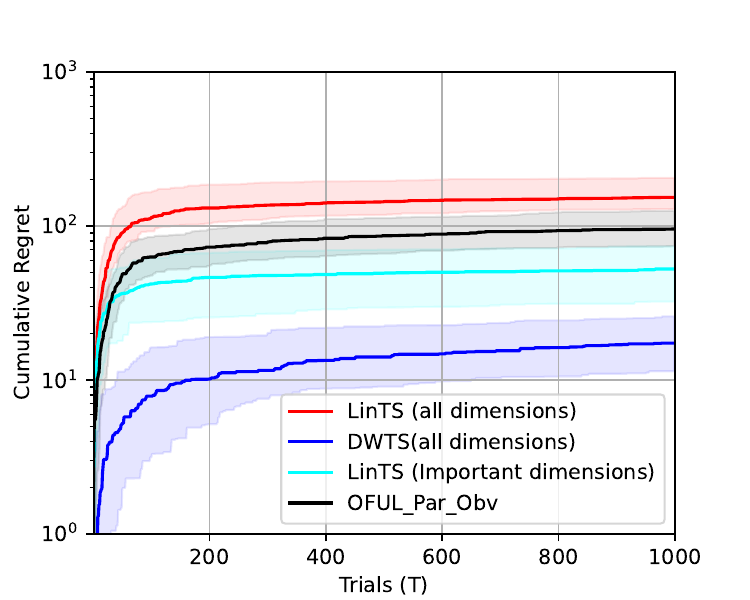}
    \caption{$p=20$}
    \label{fig:settingA}
  \end{subfigure}
  \begin{subfigure}{0.32\linewidth}
    \centering
    \includegraphics[width=\linewidth]{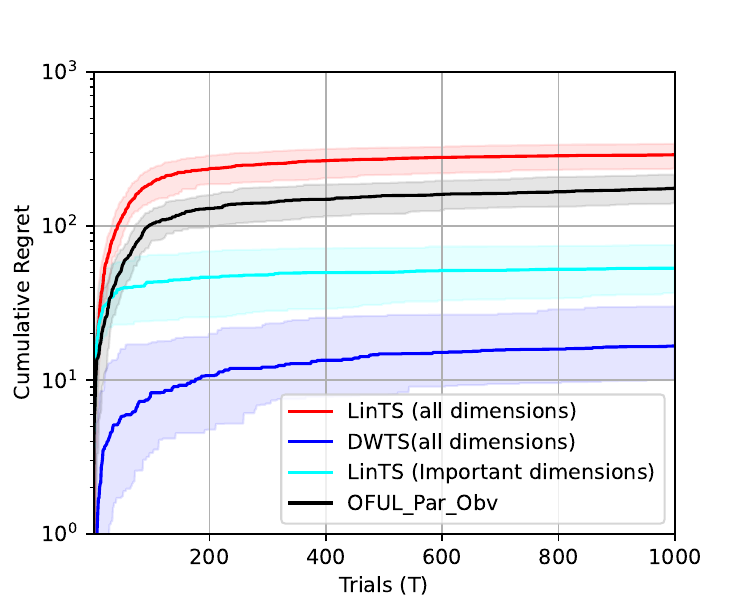}
    \caption{$p=40$}
    \label{fig:settingB}
  \end{subfigure}
  \begin{subfigure}{0.32\linewidth}
    \centering
    \includegraphics[width=\linewidth]{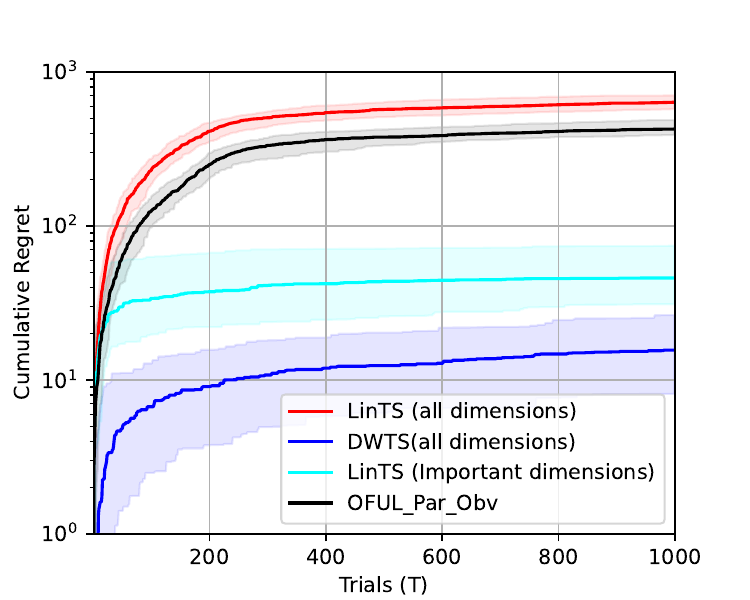}
    \caption{$p=100$}
    \label{fig:settingC}
  \end{subfigure}
  \caption{Median cumulative regret (solid) with 10\%–90\% quantile bands (shaded) over $T=1000$ rounds and across $50$ replications. \textbf{Red:} LinTS on all $p+q$ dims. \textbf{Cyan:} LinTS on $p_{\text{true}}+q$ dims. \textbf{Blue:} DWTS warm‑start on $p+q$ dims. \textbf{Black:} OFUL with Partially Observable Offline data~\cite{tennenholtz2021bandits}.} 
  \label{fig:synthetic}
\end{figure}

We report cumulative regret, summarizing the median and the 10\%–90\% quantile band over the $50$ replications. Figure~\ref{fig:synthetic} shows that DWTS (blue) attains significantly lower cumulative regret than LinTS (red) in the sparse regime and LinTS with true dimensions (cyan), and OFUL with partially observable offline data algorithm (black), thereby validating the benefits of effective dimension reduction and deconfounded warm-start. All experiments were run on an Apple M1 chip and 16 GB of RAM, without GPU acceleration. Our method runs efficiently on standard hardware with no GPU acceleration.

\section{Case Study: Patients at High Risks for Cardiovascular Events}

\subsection{Description of Clinical Data}

In this section, we evaluate the effectiveness of our proposed algorithm in the context of personalized medicine. Specifically, we used a granular dataset in our case study: the National Health and Nutrition Examination Survey (NHANES). NHANES dataset is a program of studies conducted by the National Center for Health Statistics (NCHS), a part of the Centers for Disease Control and Prevention (CDC) in the United States. NHANES is designed to assess the health and nutritional status of adults and children in the U.S. through a combination of interviews, physical examinations, and laboratory tests. 
This dataset annually gathers health and nutritional data from a nationally representative sample of about 5,000 individuals across the United States. 

For our case study, we constructed the dataset using individual-level data from NHANES, covering the period from 2009 to 2018. The dataset was organized into four distinct feature groups:
(i) Demographic features, including age, sex, and race/ethnicity;
(ii) Clinical conditions, such as current tobacco use, history of cardiovascular disease, systolic and diastolic blood pressure, body mass index, and heart rate;
(iii) Biomarkers, including hemoglobin A1c, total cholesterol, high-density lipoprotein (HDL), low-density lipoprotein (LDL), triglycerides, fasting glucose, alanine aminotransferase, potassium, serum creatinine, and urine creatinine;
and (iv) Prescribed medications, including the use of blood pressure-lowering medications (BPRx) and statins. See Table \ref{tab:data_statistics} for the details of our real-world dataset in the Appedix~\ref{sec:App}. We also focused on patients who are at a high risk for the cardiovascular events. 

The estimation of risk of cardiovascular disease (CVD) is a critical tool to predict an individual's 10-year risk of developing the first hard CVD event, including non-fatal myocardial infarction, death from coronary heart disease, or stroke. The American College of Cardiology (ACC) \citep{basu2017development} and American Heart Association (AHA) Pooled Cohort Risk Equations \citep{goff20142013} are widely used for this purpose, utilizing patient-specific factors such as age, sex, race, cholesterol levels, blood pressure, smoking status, and diabetes. 
%Estimating ASCVD risk is essential for identifying patients at high risk who may benefit from preventive strategies, such as initiating statin therapy, antihypertensive treatment, or lifestyle interventions. 
Physicians use these risk estimates to engage in shared decision-making with patients, helping to tailor treatment plans based on individualized risk profiles and to prioritize interventions that can most effectively reduce the risk of cardiovascular events. Our dataset includes CVD risk estimation from both CVD risk calculators of \cite{basu2017development} and \cite{goff20142013}.

\subsection{Creating Virtual Environment using NHANES Dataset}\label{sec:Virtual}

We construct a realistic bandit environment directly from the NHANES dataset by treating each patient record as a context vector ${X} = [Z; H]$, where $Z$ denotes observed covariates (e.g. laboratory measurements and clinical conditions) and $H$ represents latent or hidden factors such as demographics.  At each decision round, we uniformly sample an index $i$ and extract the full context $\mathbf{x}_i$.  We then compute the baseline caridiovascular disease risk score termed as $\text{CVDRisk}_i$ using standard risk calculators (\citealt{basu2017development} and \citealt{goff20142013}).  Upon selecting an action or treatment $a \in \{\text{NoRx}, \text{BPRx}\}$, we model treatment efficacy $\widetilde{r}_i(a)$ by applying an arm-specific multiplicative reduction factor $\rho_a$, that is
\(
r_i(a) = \rho_a \cdot \text{CVDRisk}_i.
\)
To capture inter-patient variability, we add Gaussian noise to compute 
\(
 r_i(a) =\widetilde{r}_i(a) + \varepsilon, \quad \varepsilon \sim \mathcal{N}(0,\sigma^2).
\)
This procedure yields a stochastic reward signal for each context–action pair, enabling rigorous evaluation of bandit algorithms in a fully data-driven, virtual clinical trial environment.

\subsection{Experimental Results on NHANES-Derived Environment}

We evaluate our proposed DWTS algorithm in a semi-synthetic environment described in Section~\ref{sec:Virtual}, constructed from the NHANES dataset. We fix the treatment space as $\mathcal{A}= \{\texttt{NoRx}, \texttt{BPRx}\}$, with $\rho_{\text{NoRx}}= 1$ representing no reduction in risk without medication, and $\rho_{\text{BPRx}}= 0.4$ to simulate the effect of blood pressure medication. 

We compare DWTS against two baselines: standard LinTS, which learns over all available covariates (including demographic, clinical, and biomarker features), and a LinTS variant restricted to only those features used in computing $\text{CVDRisk}_i$. In the DWTS experiment, we fix $Z$ to represent the clinical and biomarker features, and $H$ to denote the demographic features. For the offline phase of DWTS, we use a subset of the NHANES dataset consisting of samples with fixed (according to some non-randomized offline policy) treatment assignment and corresponding $\text{CVDRisk}_i$ values to construct the offline dataset. On this dataset, we apply DDL with a threshold of $\kappa_o = 0.01$ to identify a sparse subset of reliable covariates and to estimate deconfounded means and standard errors. In the online phase, DWTS priors are initialized using the offline mean and variance estimates of the selected covariates, while the unobserved (hidden) dimensions are handled using uninformative priors.

\begin{figure}[h]
  \centering  \includegraphics[width=0.5\linewidth]{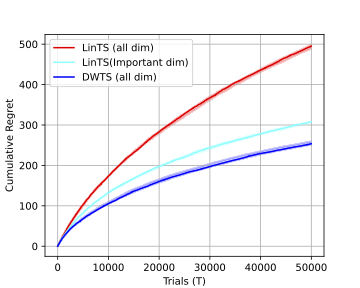}
  \caption{Median cumulative regret (solid) with 10\%–90\% quantile bands (shaded) over $T=50000$ rounds and across $10$ replications. \textbf{Red:} LinTS on all $p+q$ dims. \textbf{Cyan:} LinTS on $p_{\text{true}}+q$ dims. \textbf{Blue:} DWTS warm‑start on $p+q$ dims.}
  \label{fig:NHANES}
\end{figure}

All methods are evaluated over multiple random trajectories, and we report the median cumulative regret over time. DWTS consistently achieves lower regret, especially in early rounds, highlighting the effectiveness of incorporating causal insights from observational data into adaptive decision-making. In Figure~\ref{fig:NHANES}, we show that DWTS consistently outperforms baseline methods in cumulative regret, highlighting the value of integrating offline causal knowledge with online learning. The improvement is particularly notable in the early rounds, indicating more efficient learning and better initial decision-making. These findings suggest that DWTS offers a principled way to leverage observational medical data to accelerate and improve adaptive clinical trials.

\section{Conclusion and Limitations}

We introduced Deconfounded Warm-Start Thompson Sampling (DWTS), a novel framework for integrating confounded observational data into online decision-making, with particular relevance to adaptive clinical trials. DWTS enhances the classical LinTS algorithm by incorporating a doubly debiased feature selection procedure to identify a sparse set of reliable covariates. These covariates are then used to warm-start exploration in a reduced and deconfounded context space, significantly improving sample efficiency and policy performance in both synthetic and real-world settings.

Despite its promising results, DWTS presents several limitations that open directions for future research. First, our framework assumes that all hidden confounders become observable in the online phase -- a reasonable assumption in simulated environments, but a potentially unrealistic one in real-world clinical applications where certain factors remain unmeasured. Second, DWTS is currently restricted to linear models; extending the framework to handle non-linear outcome models or deep representations could broaden its applicability. Third, we assume a static, pre-collected observational dataset. In dynamic settings like healthcare, where patient populations and treatment protocols evolve over time, future work must address how to adapt DWTS to handle non-stationary observational data. Additionally, the lack of formal regret guarantees in this work limits theoretical understanding, particularly in the presence of partial observability. Finally, real-world decision-makers often face multiple heterogeneous offline datasets; developing principled methods for combining them remains an open challenge.

Overall, DWTS offers a principled and practical approach for leveraging observational data to accelerate online learning and improve outcomes in high-stakes decision-making domains, while motivating several important avenues for future exploration.

\bibliographystyle{abbrvnat} 
\bibliography{references}

\begin{thebibliography}{47}
\providecommand{\natexlab}[1]{#1}
\providecommand{\url}[1]{\texttt{#1}}
\expandafter\ifx\csname urlstyle\endcsname\relax
  \providecommand{\doi}[1]{doi: #1}\else
  \providecommand{\doi}{doi: \begingroup \urlstyle{rm}\Url}\fi

\bibitem[Agrawal and Goyal(2013)]{agrawal2013thompson}
S.~Agrawal and N.~Goyal.
\newblock Thompson sampling for contextual bandits with linear payoffs.
\newblock In \emph{International conference on machine learning}, pages 127--135. PMLR, 2013.

\bibitem[Banerjee et~al.(2022)Banerjee, Sinclair, Tambe, Xu, and Yu]{banerjee2022artificial}
S.~Banerjee, S.~R. Sinclair, M.~Tambe, L.~Xu, and C.~L. Yu.
\newblock Artificial replay: a meta-algorithm for harnessing historical data in bandits.
\newblock \emph{arXiv preprint arXiv:2210.00025}, 2022.

\bibitem[Basu et~al.(2017)Basu, Sussman, Berkowitz, Hayward, and Yudkin]{basu2017development}
S.~Basu, J.~B. Sussman, S.~A. Berkowitz, R.~A. Hayward, and J.~S. Yudkin.
\newblock Development and validation of risk equations for complications of type 2 diabetes (recode) using individual participant data from randomised trials.
\newblock \emph{The lancet Diabetes \& endocrinology}, 5\penalty0 (10):\penalty0 788--798, 2017.

\bibitem[Belloni et~al.(2013)Belloni, Chernozhukov, and Hansen]{belloni2013inference}
A.~Belloni, V.~Chernozhukov, and C.~Hansen.
\newblock Inference for high-dimensional sparse econometric models.
\newblock In \emph{Advances in economics and econometrics: Tenth world congress}, volume~3, pages 245--95. Cambridge University Press Cambridge, 2013.

\bibitem[Brookhart et~al.(2010)Brookhart, St{\"u}rmer, Glynn, Rassen, and Schneeweiss]{brookhart2010confounding}
M.~A. Brookhart, T.~St{\"u}rmer, R.~J. Glynn, J.~Rassen, and S.~Schneeweiss.
\newblock Confounding control in healthcare database research: challenges and potential approaches.
\newblock \emph{Medical care}, 48\penalty0 (6):\penalty0 S114--S120, 2010.

\bibitem[Buehlmann(2006)]{buehlmann2006boosting}
P.~Buehlmann.
\newblock Boosting for high-dimensional linear models.
\newblock 2006.

\bibitem[Cheng and Cai(2021)]{cheng2021adaptive}
D.~Cheng and T.~Cai.
\newblock Adaptive combination of randomized and observational data.
\newblock \emph{arXiv preprint arXiv:2111.15012}, 2021.

\bibitem[Curth and Van~der Schaar(2021)]{curth2021nonparametric}
A.~Curth and M.~Van~der Schaar.
\newblock Nonparametric estimation of heterogeneous treatment effects: From theory to learning algorithms.
\newblock In \emph{International Conference on Artificial Intelligence and Statistics}, pages 1810--1818. PMLR, 2021.

\bibitem[Dahabreh and Bibbins-Domingo(2024)]{dahabreh2024causal}
I.~J. Dahabreh and K.~Bibbins-Domingo.
\newblock Causal inference about the effects of interventions from observational studies in medical journals.
\newblock \emph{Jama}, 331\penalty0 (21):\penalty0 1845--1853, 2024.

\bibitem[Denton(2018)]{denton2018optimization}
B.~T. Denton.
\newblock Optimization of sequential decision making for chronic diseases: From data to decisions.
\newblock In \emph{Recent Advances in Optimization and Modeling of Contemporary Problems}, pages 316--348. INFORMS, 2018.

\bibitem[Efron et~al.(2004)Efron, Hastie, Johnstone, and Tibshirani]{efron2004least}
B.~Efron, T.~Hastie, I.~Johnstone, and R.~Tibshirani.
\newblock Least angle regression.
\newblock 2004.

\bibitem[Goff et~al.(2014)Goff, Lloyd-Jones, Bennett, Coady, D’agostino, Gibbons, Greenland, Lackland, Levy, O’donnell, et~al.]{goff20142013}
D.~C. Goff, D.~M. Lloyd-Jones, G.~Bennett, S.~Coady, R.~B. D’agostino, R.~Gibbons, P.~Greenland, D.~T. Lackland, D.~Levy, C.~J. O’donnell, et~al.
\newblock 2013 acc/aha guideline on the assessment of cardiovascular risk: a report of the american college of cardiology/american heart association task force on practice guidelines.
\newblock \emph{Journal of the American College of Cardiology}, 63\penalty0 (25 Part B):\penalty0 2935--2959, 2014.

\bibitem[Guo et~al.(2022)Guo, {\'C}evid, and B{\"u}hlmann]{guo2022doubly}
Z.~Guo, D.~{\'C}evid, and P.~B{\"u}hlmann.
\newblock Doubly debiased lasso: High-dimensional inference under hidden confounding.
\newblock \emph{Annals of statistics}, 50\penalty0 (3):\penalty0 1320, 2022.

\bibitem[Hao et~al.(2023)Hao, Jain, Lattimore, Van~Roy, and Wen]{hao2023leveraging}
B.~Hao, R.~Jain, T.~Lattimore, B.~Van~Roy, and Z.~Wen.
\newblock Leveraging demonstrations to improve online learning: Quality matters.
\newblock In \emph{International Conference on Machine Learning}, pages 12527--12545. PMLR, 2023.

\bibitem[Hatt and Feuerriegel(2021)]{hatt2021estimating}
T.~Hatt and S.~Feuerriegel.
\newblock Estimating average treatment effects via orthogonal regularization.
\newblock In \emph{Proceedings of the 30th ACM International Conference on Information \& Knowledge Management}, pages 680--689, 2021.

\bibitem[Hatt et~al.(2022)Hatt, Berrevoets, Curth, Feuerriegel, and van~der Schaar]{hatt2022combining}
T.~Hatt, J.~Berrevoets, A.~Curth, S.~Feuerriegel, and M.~van~der Schaar.
\newblock Combining observational and randomized data for estimating heterogeneous treatment effects.
\newblock \emph{arXiv preprint arXiv:2202.12891}, 2022.

\bibitem[Hu et~al.(2021)Hu, Ji, and Li]{hu2021estimating}
L.~Hu, J.~Ji, and F.~Li.
\newblock Estimating heterogeneous survival treatment effect in observational data using machine learning.
\newblock \emph{Statistics in medicine}, 40\penalty0 (21):\penalty0 4691--4713, 2021.

\bibitem[Ilse et~al.(2021)Ilse, Forr{\'e}, Welling, and Mooij]{ilse2021combining}
M.~Ilse, P.~Forr{\'e}, M.~Welling, and J.~M. Mooij.
\newblock Combining interventional and observational data using causal reductions.
\newblock \emph{arXiv preprint arXiv:2103.04786}, 2021.

\bibitem[Kallus et~al.(2018)Kallus, Puli, and Shalit]{kallus2018removing}
N.~Kallus, A.~M. Puli, and U.~Shalit.
\newblock Removing hidden confounding by experimental grounding.
\newblock \emph{Advances in neural information processing systems}, 31, 2018.

\bibitem[Keyvanshokooh et~al.(2025)Keyvanshokooh, Zhalechian, Shi, Van~Oyen, and Kazemian]{keyvanshokooh2025contextual}
E.~Keyvanshokooh, M.~Zhalechian, C.~Shi, M.~P. Van~Oyen, and P.~Kazemian.
\newblock Contextual learning with online convex optimization: Theory and application to medical decision-making.
\newblock \emph{Management Science}, 2025.

\bibitem[Kim et~al.(2025)Kim, Park, Iyengar, Zeevi, and Oh]{kim2025linear}
W.~Kim, S.~Park, G.~Iyengar, A.~Zeevi, and M.-h. Oh.
\newblock Linear bandits with partially observable features.
\newblock \emph{arXiv preprint arXiv:2502.06142}, 2025.

\bibitem[Kim et~al.(2019)Kim, Hao, Mallavarapu, Park, and Kang]{kim2019hi}
Y.~Kim, J.~Hao, T.~Mallavarapu, J.~Park, and M.~Kang.
\newblock Hi-lasso: High-dimensional lasso.
\newblock \emph{IEEE Access}, 7:\penalty0 44562--44573, 2019.

\bibitem[K{\"u}nzel et~al.(2019)K{\"u}nzel, Sekhon, Bickel, and Yu]{kunzel2019metalearners}
S.~R. K{\"u}nzel, J.~S. Sekhon, P.~J. Bickel, and B.~Yu.
\newblock Metalearners for estimating heterogeneous treatment effects using machine learning.
\newblock \emph{Proceedings of the national academy of sciences}, 116\penalty0 (10):\penalty0 4156--4165, 2019.

\bibitem[Kuzmanovic et~al.(2021)Kuzmanovic, Hatt, and Feuerriegel]{kuzmanovic2021deconfounding}
M.~Kuzmanovic, T.~Hatt, and S.~Feuerriegel.
\newblock Deconfounding temporal autoencoder: estimating treatment effects over time using noisy proxies.
\newblock In \emph{Machine Learning for Health}, pages 143--155. PMLR, 2021.

\bibitem[Lattimore and Szepesv{\'a}ri(2020)]{lattimore2020bandit}
T.~Lattimore and C.~Szepesv{\'a}ri.
\newblock \emph{Bandit algorithms}.
\newblock Cambridge University Press, 2020.

\bibitem[Louizos et~al.(2017)Louizos, Shalit, Mooij, Sontag, Zemel, and Welling]{louizos2017causal}
C.~Louizos, U.~Shalit, J.~M. Mooij, D.~Sontag, R.~Zemel, and M.~Welling.
\newblock Causal effect inference with deep latent-variable models.
\newblock \emph{Advances in neural information processing systems}, 30, 2017.

\bibitem[Luo and Chen(2014)]{luo2014sequential}
S.~Luo and Z.~Chen.
\newblock Sequential lasso cum ebic for feature selection with ultra-high dimensional feature space.
\newblock \emph{Journal of the American Statistical Association}, 109\penalty0 (507):\penalty0 1229--1240, 2014.

\bibitem[Mei and Shi(2024)]{mei2024lasso}
Z.~Mei and Z.~Shi.
\newblock On lasso for high dimensional predictive regression.
\newblock \emph{Journal of Econometrics}, 242\penalty0 (2):\penalty0 105809, 2024.

\bibitem[N{\o}rgaard et~al.(2017)N{\o}rgaard, Ehrenstein, and Vandenbroucke]{norgaard2017confounding}
M.~N{\o}rgaard, V.~Ehrenstein, and J.~P. Vandenbroucke.
\newblock Confounding in observational studies based on large health care databases: problems and potential solutions--a primer for the clinician.
\newblock \emph{Clinical epidemiology}, pages 185--193, 2017.

\bibitem[Pearl(2009)]{pearl2009causality}
J.~Pearl.
\newblock \emph{Causality}.
\newblock Cambridge university press, 2009.

\bibitem[Powell et~al.(2022)Powell, Clark, Alyakin, Vogelstein, and Hart]{powell2022exploration}
M.~Powell, C.~Clark, A.~Alyakin, J.~T. Vogelstein, and B.~Hart.
\newblock Exploration of residual confounding in analyses of associations of metformin use and outcomes in adults with type 2 diabetes.
\newblock \emph{JAMA Network Open}, 5\penalty0 (11):\penalty0 e2241505--e2241505, 2022.

\bibitem[Rosenman et~al.(2023)Rosenman, Basse, Owen, and Baiocchi]{rosenman2023combining}
E.~T. Rosenman, G.~Basse, A.~B. Owen, and M.~Baiocchi.
\newblock Combining observational and experimental datasets using shrinkage estimators.
\newblock \emph{Biometrics}, 79\penalty0 (4):\penalty0 2961--2973, 2023.

\bibitem[Schneeweiss et~al.(2017)Schneeweiss, Eddings, Glynn, Patorno, Rassen, and Franklin]{schneeweiss2017variable}
S.~Schneeweiss, W.~Eddings, R.~J. Glynn, E.~Patorno, J.~Rassen, and J.~M. Franklin.
\newblock Variable selection for confounding adjustment in high-dimensional covariate spaces when analyzing healthcare databases.
\newblock \emph{Epidemiology}, 28\penalty0 (2):\penalty0 237--248, 2017.

\bibitem[Sharma et~al.(2020)Sharma, Basu, Shanmugam, and Shakkottai]{sharma2020under}
N.~Sharma, S.~Basu, K.~Shanmugam, and S.~Shakkottai.
\newblock On under-exploration in bandits with mean bounds from confounded data.
\newblock \emph{arXiv preprint arXiv:2002.08405}, 2020.

\bibitem[Shi et~al.(2019)Shi, Blei, and Veitch]{shi2019adapting}
C.~Shi, D.~Blei, and V.~Veitch.
\newblock Adapting neural networks for the estimation of treatment effects.
\newblock \emph{Advances in neural information processing systems}, 32, 2019.

\bibitem[Stuart et~al.(2011)Stuart, Cole, Bradshaw, and Leaf]{stuart2011use}
E.~A. Stuart, S.~R. Cole, C.~P. Bradshaw, and P.~J. Leaf.
\newblock The use of propensity scores to assess the generalizability of results from randomized trials.
\newblock \emph{Journal of the Royal Statistical Society Series A: Statistics in Society}, 174\penalty0 (2):\penalty0 369--386, 2011.

\bibitem[Stukel et~al.(2007)Stukel, Fisher, Wennberg, Alter, Gottlieb, and Vermeulen]{stukel2007analysis}
T.~A. Stukel, E.~S. Fisher, D.~E. Wennberg, D.~A. Alter, D.~J. Gottlieb, and M.~J. Vermeulen.
\newblock Analysis of observational studies in the presence of treatment selection bias: effects of invasive cardiac management on ami survival using propensity score and instrumental variable methods.
\newblock \emph{Jama}, 297\penalty0 (3):\penalty0 278--285, 2007.

\bibitem[Tang et~al.(2023)Tang, Jain, Hao, and Wen]{tang2023efficient}
D.~Tang, R.~Jain, B.~Hao, and Z.~Wen.
\newblock Efficient online learning with offline datasets for infinite horizon mdps: A bayesian approach.
\newblock \emph{arXiv preprint arXiv:2310.11531}, 2023.

\bibitem[Tennenholtz et~al.(2021)Tennenholtz, Shalit, Mannor, and Efroni]{tennenholtz2021bandits}
G.~Tennenholtz, U.~Shalit, S.~Mannor, and Y.~Efroni.
\newblock Bandits with partially observable confounded data.
\newblock In \emph{Uncertainty in Artificial Intelligence}, pages 430--439. PMLR, 2021.

\bibitem[Tibshirani(1996)]{tibshirani1996regression}
R.~Tibshirani.
\newblock Regression shrinkage and selection via the lasso.
\newblock \emph{Journal of the Royal Statistical Society Series B: Statistical Methodology}, 58\penalty0 (1):\penalty0 267--288, 1996.

\bibitem[Wagenmaker and Pacchiano(2023)]{wagenmaker2023leveraging}
A.~Wagenmaker and A.~Pacchiano.
\newblock Leveraging offline data in online reinforcement learning.
\newblock In \emph{International Conference on Machine Learning}, pages 35300--35338. PMLR, 2023.

\bibitem[Wager and Athey(2018)]{wager2018estimation}
S.~Wager and S.~Athey.
\newblock Estimation and inference of heterogeneous treatment effects using random forests.
\newblock \emph{Journal of the American Statistical Association}, 113\penalty0 (523):\penalty0 1228--1242, 2018.

\bibitem[Wang and Blei(2019)]{wang2019blessings}
Y.~Wang and D.~M. Blei.
\newblock The blessings of multiple causes.
\newblock \emph{Journal of the American Statistical Association}, 114\penalty0 (528):\penalty0 1574--1596, 2019.

\bibitem[Yamada et~al.(2014)Yamada, Jitkrittum, Sigal, Xing, and Sugiyama]{yamada2014high}
M.~Yamada, W.~Jitkrittum, L.~Sigal, E.~P. Xing, and M.~Sugiyama.
\newblock High-dimensional feature selection by feature-wise kernelized lasso.
\newblock \emph{Neural computation}, 26\penalty0 (1):\penalty0 185--207, 2014.

\bibitem[Yang and Ding(2019)]{yang2019combining}
S.~Yang and P.~Ding.
\newblock Combining multiple observational data sources to estimate causal effects.
\newblock \emph{Journal of the American Statistical Association}, 2019.

\bibitem[Yao et~al.(2018)Yao, Li, Li, Huai, Gao, and Zhang]{yao2018representation}
L.~Yao, S.~Li, Y.~Li, M.~Huai, J.~Gao, and A.~Zhang.
\newblock Representation learning for treatment effect estimation from observational data.
\newblock \emph{Advances in neural information processing systems}, 31, 2018.

\bibitem[Zhang et~al.(2019)Zhang, Agarwal, Daum{\'e}~III, Langford, and Negahban]{zhang2019warm}
C.~Zhang, A.~Agarwal, H.~Daum{\'e}~III, J.~Langford, and S.~N. Negahban.
\newblock Warm-starting contextual bandits: Robustly combining supervised and bandit feedback.
\newblock \emph{arXiv preprint arXiv:1901.00301}, 2019.

\end{thebibliography}

\appendix

\section{Appendix}~\label{sec:App}

\begin{table}[htbp]
\centering
\caption{Summary of Patient Characteristics in the NHANES dataset: Mean Values with Standard Deviations and Counts with Percentages. The standard deviations and percentages are summarized in parentheses.}
%\begin{adjustbox}%{width=0.65\linewidth}
\resizebox{0.6\textwidth}{!}{
\begin{tabular}{lrc}\hline\hline
& \multicolumn{2}{c}{NHANES}     \\ 
& \multicolumn{2}{c}{$(n=28,409)$}  \\\hline
\multicolumn{3}{l}{\textbf{Demographic}}  \\ \hline 
Age, year                              & $45.99$   & $(17.24)$      \\
\multicolumn{3}{l}{{Sex}} \\
\quad Female                            & $14,610$   &$(51.42\%)$    \\
\quad Male                              & $13,799$   &$(48.58\%)$   \\
\multicolumn{3}{l}{{Race and Ethnicity}} \\ 
\quad Black, non-Hispanic             & $6,367$    &$(22.41\%)$     \\
\quad White, non-Hispanic             & $10,377$   &$(36.38\%)$     \\
\quad Hispanic                        & $7,480$    &$(26.32\%)$     \\
\quad Other                           & $4,225$    &$(14.89\%)$     \\
\hline
\multicolumn{3}{l}{\textbf{Clinical Conditions}}  \\ 
\hline 
Tobacco smoking, current        & $14,789$      &$(52.05\%)$ \\
History of Cardiovascular Disease   &$1,667$    &$(5.86\%)$  \\
\multicolumn{3}{l}{{Blood Pressure, seated}} \\
\quad Systolic, mm Hg                   &$122.98$   &$(18.11)$ \\
\quad Diastolic, mm Hg                  &$70.47$    &$(12.77)$ \\
BMI, kg/$\text{m}^2$                    &$29.19$    &$(7.21)$  \\
Heart Rate, bpm                         &$79.64$    &$(4.94)$  \\ 
\hline
\multicolumn{3}{l}{\textbf{Biomarkers}}  \\ 
\hline
HBA1c, $\%$                             &$5.75$      &$(1.09)$  \\
Total Cholesterol, mg/dL                &$189.93$   &$(41.44)$  \\
HDL Cholesterol, mg/dL                  &$52.81$    &$(15.86)$  \\
LDL Cholesterol, mg/dL                  &$114.46$   &$(42.27)$  \\
Triglycerides, mg/dL                    &$113.61$   &$(102.61)$ \\
Fasting Plasma Glucose, mg/dL           &$108.29$   &$(34.78)$  \\
Alanine Aminotransferase, IU/L          &$24.83$    &$(20.05)$  \\
Potassium, mmol/L                       &$3.98$     &$(0.34)$   \\
Serum Creatinine, mg/dL                 &$0.88$     &$(0.44)$   \\
Urine Creatinine, md/dL                 &$127.23$   &$(83.75)$  \\
Urine Albumin:Creatinine ratio, mg/g    &$46.96$    &$(648.48)$ \\
\hline
\multicolumn{3}{l}{\textbf{Prescribed Medications}}  \\ 
\hline
BPRx        &$7,823$ &$(27.53\%)$ \\ 
Statins     &$4,683$ &$(16.48\%)$ \\ 
\hline
\end{tabular}
}
\label{tab:data_statistics}
\end{table}

\end{document}